\documentclass{article}

\usepackage{microtype}
\usepackage{graphicx}
\usepackage{subfigure}
\usepackage{booktabs} 
\usepackage{amssymb}
\usepackage{amsmath}
\usepackage{float} 
\usepackage{multirow}

\usepackage{hyperref}

\usepackage[accepted]{icml2020} 

\icmltitlerunning{Hierarchical Protein Function Prediction with Tail-GNNs}

\begin{document}

\twocolumn[
\icmltitle{Hierarchical Protein Function Prediction with Tail-GNNs}



\icmlsetsymbol{equal}{*}

\begin{icmlauthorlist}
\icmlauthor{Stefan Spalevi\'{c}}{matf}
\icmlauthor{Petar Veli\v{c}kovi\'{c}}{dm}
\icmlauthor{Jovana Kova\v{c}evi\'{c}}{matf}
\icmlauthor{Mladen Nikoli\'{c}}{matf}
\end{icmlauthorlist}

\icmlaffiliation{matf}{Faculty of Mathematics, University of Belgrade}
\icmlaffiliation{dm}{DeepMind}

\icmlcorrespondingauthor{Stefan Spalevi\'{c}}{spalemon94@gmail.com}

\icmlkeywords{Machine Learning, ICML}

\vskip 0.3in
]



\printAffiliationsAndNotice{} 

\begin{abstract}
Protein function prediction may be framed as predicting subgraphs (with certain closure properties) of a directed acyclic graph describing the hierarchy of protein functions. Graph neural networks (GNNs), with their built-in inductive bias for relational data, are hence naturally suited for this task. However, in contrast with most GNN applications, the graph is not related to the input, but to the \emph{label} space. Accordingly, we propose \textbf{Tail-GNNs}, neural networks which naturally compose with the output space of any neural network for multi-task prediction, to provide relationally-reinforced labels. For protein function prediction, we combine a Tail-GNN with a dilated convolutional network which learns representations of the protein sequence, making significant improvement in $F_1$ score and demonstrating the ability of Tail-GNNs to learn useful representations of labels and exploit them in real-world problem solving.
\end{abstract}


\section{Introduction}

Knowing the function of a protein informs us on its biological role in the organism. With large numbers of genomes being sequenced every year, there is a rapidly growing number of newly discovered proteins. Protein function is most reliably determined in \emph{wet lab} experiments, but current experimental methods are too slow for such quick income of novel proteins. Therefore, the development of tools for automated  prediction of protein functions is necessary. Fast and accurate prediction of protein function is especially important in the context of human diseases since many of them are associated with specific protein functions.

\begin{figure*}
    \includegraphics[width=\linewidth, scale = 0.30]{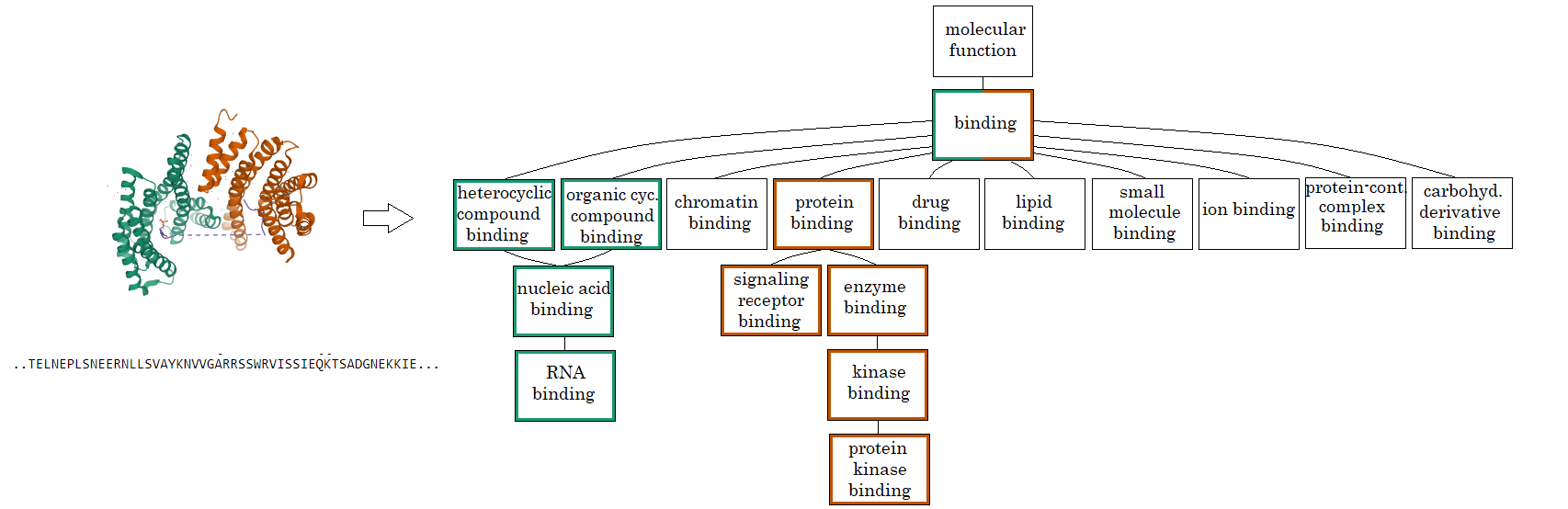}
    \caption{Representation of a function subgraph on a small subset of the ontology we leveraged. Assume that the input protein has three  functions: \textit{RNA binding}, \textit{signaling receptor binding} and \textit{protein kinase binding}. Its function subgraph contains all predecessors of these functions (e.g. \textit{nucleic acid binding}, \textit{enzyme binding}, \textit{binding}). Note that, as we go deeper in the ontology, the functions associated with the nodes become more specialized.}
    \label{fig:mala_ontologija}
\end{figure*}

The space of all known protein functions is defined by a directed acyclic graph known as the \emph{Gene Ontology} (GO) \cite{ashburner2000gene}, where each node represents one function and each edge encodes a hierarchical relationship between two functions, such as {\em is-a} or {\em part-of} (refer to Figure \ref{fig:mala_ontologija} for a visualisation). For every protein, its functions constitute a subgraph of GO, consistent in the sense that it is closed with respect to the predecessor relationship. GO contains thousands of nodes, with function subgraphs usually having dozens of nodes for each protein. Hence, the output of the protein function prediction problem is a \emph{subgraph} of a hierarchically-structured graph.


This opens up a clear path of application for graph representation learning \cite{bronstein2017geometric,hamilton2017representation,battaglia2018relational}, especially \emph{graph neural networks} (GNNs) \cite{kipf2016semi,velivckovic2017graph,gilmer2017neural,corso2020principal}, given their natural inductive bias towards processing relational data. 

One key aspect in which the protein function prediction task differs from most applications of graph representation learning, however, is in the fact that the graph is specified in the \emph{label} space---that is, we are given a multilabel classification task in which we have known relational inductive biases over the individual labels (e.g. if protein $X$ has function $F$, it must also have \emph{all predecessor functions} of $F$ under the closure constraint).

Driven by the requirement for a GNN to operate in the \emph{label} space, we propose \textbf{Tail-GNN}, a graph neural network which learns representations of \emph{labels}, introducing relational inductive biases into the \emph{flat} label predictions of a feedforward neural network. Our results demonstrate that introducing this inductive bias provides significant gains on the protein function prediction task, paving the way to many other possible applications in the sciences (e.g., prediction of spatial phenomena over several correlated locations \cite{radosavljevic2010, djuric2015}, traffic state estimation \cite{djuric2011}, and polypharmacy side effect prediction \cite{zitnik2018modeling,deac2019drug}).

\section{Tail-GNNs}

\begin{figure*}
    \includegraphics[width=\linewidth]{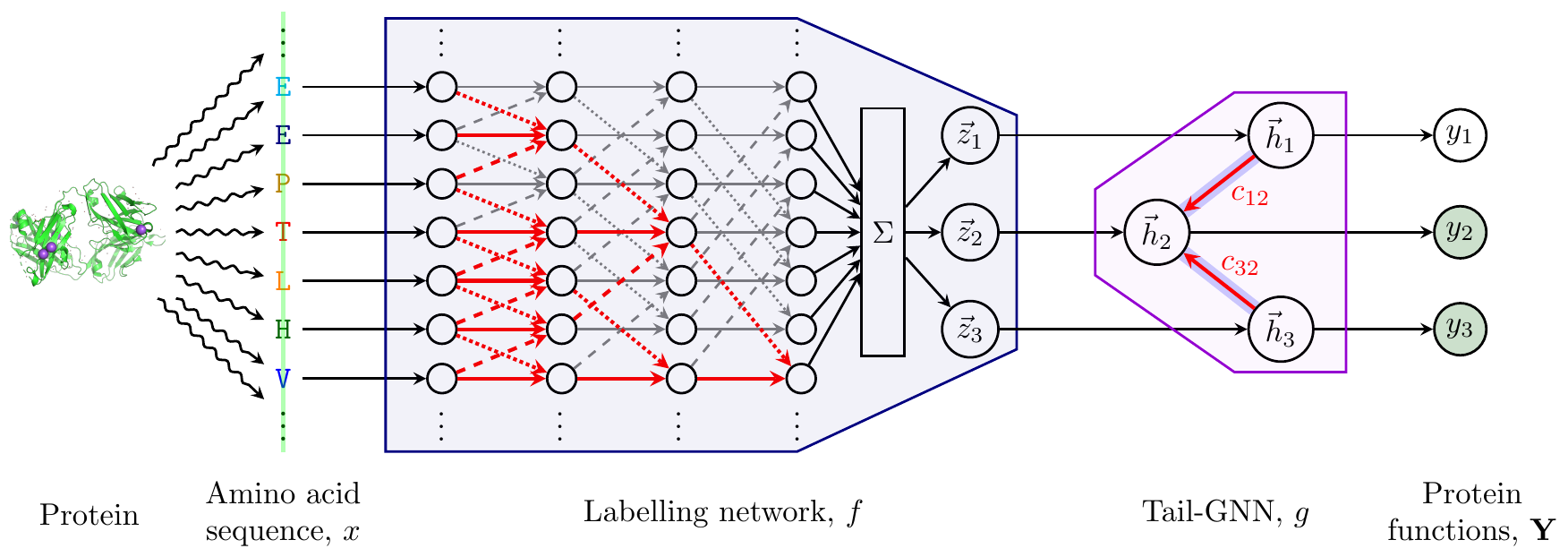}
    \caption{A high-level overview of the protein function modelling setup in this paper. Proteins are represented using their amino acid sequences ($x$), and are passed through the labelling network ($f$), to compute latent vectors for each label ($\vec{z}_i$). These latent vectors are passed to the Tail-GNN ($g$), which repeatedly aggregates their information along the edges of the gene ontology graph, computing an updated latent representation of each label ($\vec{h}_i$). Finally, a linear layer predicts the probability of the protein having the corresponding functions ($y_i$). The labelling network relies on \emph{dilated convolutions} followed by global average pooling and reshaping. Note how dilated convolutions allow for an exponentially increasing receptive field at each amino acid.}
    \label{fig:skica}
\end{figure*}

In this section, we will describe an abstract model which takes advantage of a Tail-GNN, followed by an overview and intuition for the specific architectural choices we used for the protein prediction task. The entire setup from this section may be visualised in Figure \ref{fig:skica}.

Generally, we have a multi-label prediction task, from inputs $x\in \mathcal{X}$, to outputs $y_i\in \mathcal{Y}_i$, for each label $i\in\mathbb{L}$. We are also aware that there exist relations between labels, which we explicitly encode using a binary adjacency matrix ${\bf A}\in\mathbb{R}^{|\mathbb{L}|\times|\mathbb{L}|}$, such that ${\bf A}_{ij} = 1$ implies that the prediction for label $j$ can be related\footnote{Note that different kinds of entries in ${\bf A}$ are also allowed, in case we would like to explicitly account for edge features.} with the prediction for label $i$.

Our setup consists of a {\bf labeller network}
\begin{equation}
f: \mathcal{X}\rightarrow(\mathcal{Z}_1\times\mathcal{Z}_2\times\dots\times\mathcal{Z}_{|\mathbb{L}|})
\end{equation}
which attaches \emph{latent vectors} $f(x) = {\bf Z} = \{\vec{z}_1, \dots, \vec{z}_{|\mathbb{L}|}\}$, to each label $i$, for a given input $x$. Typically, these will be $k$-dimensional real-valued vectors, i.e. $\mathcal{Z}_i = \mathbb{R}^k$.

These labels are then provided to the {\bf Tail-GNN} layer $g$, which is a node-level predictor; treating each label $i$ as a node in a graph, $\vec{z}_i$ as its corresponding node features, and ${\bf A}$ as its corresponding adjacency matrix, it produces a prediction for each node:
\begin{equation}
    g: \mathbb{R}^{|\mathbb{L}|\times k} \times \mathbb{R}^{|\mathbb{L}|\times|\mathbb{L}|}\rightarrow(\mathcal{Y}_1\times\mathcal{Y}_2\times\dots\times\mathcal{Y}_{|\mathbb{L}|})
\end{equation}
That is, $g(f(x), {\bf A}) = g({\bf Z}, {\bf A}) = {\bf Y} = (y_1, \dots, y_{|\mathbb{L}|})$, provides the final predictions for the model in each label. As implied, the {\bf Tail-GNN} is typically implemented within the graph neural network \cite{scarselli2008graph} framework, explicitly including the relational information.

Assuming $f$ and $g$ are differentiable w.r.t. their parameters, the entire system can be end-to-end optimised via gradient descent on the label errors w.r.t. ground-truth values.

In our specific case, the inputs $x$ are protein sequences of one-hot encoded amino acids, and outputs $y_i$ are binary labels indicating presence or absence of individual functions for those proteins.

Echoing the protein modelling results of Fast-Parapred \cite{deac2019attentive}, we have used a deep \emph{dilated} convolutional neural network for $f$ (similarly as in ByteNet \cite{kalchbrenner2016neural} and WaveNet \cite{oord2016wavenet}). This architecture provides a parallelisable way of modelling amino-acid sequences without sacrificing performance compared to RNN encoders. This labelling network is \emph{fully convolutional} \cite{springenberg2014striving}: it predicts $|\mathbb{L}|\times k$ latent features for each amino acid, followed by global average pooling and reshaping the output to obtain a length-$k$ vector for each label.

As we know that the gene ontology edges encode explicit containment relations between function labels, our Tail-GNN $g$ is closely related to the GCN model \cite{kipf2016semi}. At each step, we update latent features $\vec{h}_i$ in each label by aggregating neighbourhood features across edges:
\begin{equation}
    \vec{h}'_i = \mathrm{ReLU}\left(\sum_{j\in\mathcal{N}_i}c_{ji}{\bf W}\vec{h}_j\right)
\end{equation}
where $\mathcal{N}_i$ is the one-hop neighbourhood of label $i$ in the GO, ${\bf  W}$ is a shared weight matrix parametrising a linear transformation in each node, and $c_{ji}$ is a coefficient of interaction from node $j$ to node $i$, for which we attempt several variants:  \emph{sum-pooling} \cite{xu2018powerful} ($c_{ji} = 1$), \emph{mean-pooling} \cite{hamilton2017inductive} ($c_{ji}=\frac{1}{|\mathcal{N}_i|}$), and \emph{graph attention} ($c_{ji}=a(\vec{h}_i, \vec{h}_j)$, where $a$ is an \emph{attention function} producing scalar coefficients). We use the same attention mechanism as used in GAT \cite{velivckovic2017graph}.

Lastly, we also attempt to explicitly align with the containment inductive bias by leveraging \emph{max-pooling}:
\begin{equation}
    \vec{h}'_i = \mathrm{ReLU}\left(\max_{j\in\mathcal{N}_i}{\bf W}\vec{h}_j\right)
\end{equation}
where $\max$ is performed elementwise.

The final layer of our network is a shared linear layer, followed by a logistic sigmoid activation. It takes the latent label representations produced by Tail-GNN and predicts a scalar value for each label, indicating the probability of the protein having the corresponding function. We optimise the entire network end-to-end using binary cross-entropy on the ground-truth functions.

It is interesting to note that, performing constrained relational computations in the label space, the operation of the Tail-GNN can be closely related to \emph{conditional random fields} (CRFs) \cite{laferty2001crf, Krhenbhl2011EfficientII, cuong2014conditional, Belanger2016StructuredPE, Arnab2018}. CRFs have been combined with GNNs in prior work \cite{ma2018cgnf,gao2019conditional}, primarily as a means of strengthening the GNN prediction; in our work, we express all computations using GNNs alone, relying on the fact that, if optimal, Tail-GNNs could learn to specialise to the computations of the CRF through \emph{neural execution} \cite{velivckovic2019neural}, but will in principle have an opportunity to learn more \emph{data-driven} rules for message passing between different labels. 

Further, Tail-GNNs share some similarities with \emph{gated propagation networks} (GPNs) \cite{liu2019learning}, which leverage class relations to compute \emph{class prototypes} for meta-learning \cite{snell2017prototypical}. While both GPNs and Tail-GNNs perform GNN computations over a graph in the label space, the aim of GPNs is to compute structure-informed prototypes for a 1-NN classifier, while here we focus on multi-task predictions and directly produce outputs in an end-to-end differentiable fashion.

Beyond operating in the label space, GNNs have seen prior applications to protein function modelling through explicitly taking into account either the protein's residue contact map \cite{gligorijevic2019structure} or existing protein-protein interaction (PPI) networks. Especially, \citet{hamilton2017inductive} provide the first study of explicitly running GNNs over PPI graphs in order to predict gene ontology signatures \cite{zitnik2017predicting}. However, as these models rely on an existence of either a reliable contact map or PPI graph, they cannot be reliably used to predict functions for novel proteins (for which these may not yet be known). Such information, if assumed available, may be explicitly included as a relational component within the labeller network.

\section{Experimental Evaluation}

\subsection{Dataset}

We used training sequences and functional annotations from CAFA3, a protein function prediction challenge \cite{zhou2019cafa}. The functional annotations were represented by functional terms of the hierarchical structure of the Gene Ontology (GO) \cite{ashburner2000gene}---the version released in April 2020. Out of the three large groups of functions represented in GO, we used the \emph{Molecular Function Ontology} (MFO) which contains 11,113 terms. Function subgraphs for each protein were obtained by propagating functional annotations to the root. We discarded obsolete nodes and functions occurring in less than 500 proteins in the original dataset, obtaining a reduced ontology with 123 nodes and 145 edges. Next, we eliminated proteins whose function subgraph contained only the root node (which is always active), as well as proteins longer than 1,000 amino acids. 

All of the above constraints were devised with the aim of keeping the downstream task relevant, while at the same time simpler for the dilated convolutions to model---delegating most of the subsequent representational effort to the Tail-GNN. The final dataset contains 31,243 proteins, with an average sequence length of 431 amino acids. Average number of protein functions per protein is 7.

\subsection{Training specifics}

The dataset was randomly split into training/validation/test sets, with a rough proportion of 68:17:15 percent. We counted up the individual label occurrences within these datasets, observing that the split was appropriately stratified across all of them. The time of characterization of protein function was not taken into account since the aim was to examine whether GNN method is able to cope with structural labels.

The architectural hyperparameters were determined based on the validation set performance, using the $F_1$ score---a suitable measure for imbalanced label problems, which is also commonly used for evaluating models in CAFA challenges \cite{zhou2019cafa}. Via thorough hyperparameter sweeps, we decided on a labelling network of six dilated convolutional layers, with exponentially increasing dilation rate. Initially the individual amino acids are embedded into 16 features, and the individual layers compute $\{32, 64, 128, 256, 512, 512\}$ features each, mirroring the results of \citet{deac2019attentive}.

For predicting functions directly from the labelling network, we follow with a linear layer of $123$ features and global average pooling across amino acid positions, predicting the probability of each function occurring. 

When pairing with Tail-GNN, however, the linear layer computes $123k$ features, with $k$ being the number of latent features computed per label (i.e. the dimensionality of the $\vec{z}_i$ vectors). We swept various small\footnote{Further increasing $k$ quickly leads to an increase in parameter count, leading to overfitting and memory issues.} values of $k$, finding $k=9$ to perform optimally. 

In addition, we concatenate five \emph{spectral} features to each input node to the Tail-GNN, in the form of the five eigenvectors corresponding to the five largest eigenvalues of the graph Laplacian---inspired by the Graph Fourier Transform of \citet{bruna2013spectral}. 

For each choice of Tail-GNN aggregation, we evaluated one and two GNN layers of $16$ features each, followed by a linear classifier for protein functions. We also assessed performance without incorporating the spectral features.

All models are optimising the binary cross-entropy on the function predictions using the Adam SGD optimiser \cite{kingma2014adam} (with learning rate $0.001$ and batch size of $32$), incorporating class weights to account for any imbalance. We train for $200$ epochs with early stopping on the validation $F_1$, with a patience of $20$ epochs.

\subsection{Results}

We evaluate the recovered optimised models across five random seeds. Results are given in Table~\ref{tab:evaluation}; the \emph{labelling network} is the baseline dilated convolutional network without leveraging GNNs. Additionally, we provide results across a variety of Tail-GNN configurations. Our results are consistent with the top-10 performance metrics in the CAFA3 challenge \cite{zhou2019cafa} but the direct comparison was not possible since we use a reduced ontology.


Our results demonstrate a significant performance gain associated with appending Tail-GNN to the labelling network, specifically, when using the \emph{sum} aggregator. While less aligned to the containment relation than maximisation, summation is also more ``forgiving'' with respect to any labelling mistakes: if Tail-GNN-max had learnt to perfectly implement containment, any mistakenly labelled leaves would cause large chunks of the ontology to be misclassified.

Further, we discover a performance gain associated with including the Laplacian eigenvectors: including them as node features, and a low-frequency indicator of global graph features, further improves the results of the Tail-GNN-sum.

While much of our analysis  was centered around the protein function prediction task, we conclude by noting that the way Tail-GNNs are defined is task-agnostic, and could easily see application in other areas of the sciences (as discussed in the Introduction), with minimal modification to the setup.

\begin{table}
\centering
\caption{Values of $F_1$ score on our validation and test datasets for all considered architectures, aggregated over  five random seeds.}
\begin{tabular}{lcc} \toprule 

 
{\bf Model} &   {\bf Validation} $F_1$  & {\bf Test} $F_1$ \\ \midrule 

Labelling network & $0.582 \pm 0.003$  &  $0.584 \pm 0.003$ \\ 
Tail-GNN-mean & $0.583 \pm 0.006$ & $0.586 \pm 0.004$  \\
Tail-GNN-GAT & $0.582 \pm 0.004$  &  $0.587 \pm 0.005$ \\
Tail-GNN-max & $0.581 \pm 0.002$ & $0.585 \pm 0.004$  \\
Tail-GNN-sum & ${\bf 0.596} \pm 0.003$  &  ${\bf  0.600} \pm 0.003$ \\ \midrule
Tail-GNN-sum &  \multirow{2}{*}{$0.587 \pm 0.007$} &  \multirow{2}{*}{$0.590 \pm 0.008$} \\ 
(no spectral fts.)\\
\bottomrule
\end{tabular}
\label{tab:evaluation}

\end{table}




\bibliographystyle{icml2020}
\bibliography{example_paper}

\begin{thebibliography}{35}
\providecommand{\natexlab}[1]{#1}
\providecommand{\url}[1]{\texttt{#1}}
\expandafter\ifx\csname urlstyle\endcsname\relax
  \providecommand{\doi}[1]{doi: #1}\else
  \providecommand{\doi}{doi: \begingroup \urlstyle{rm}\Url}\fi

\bibitem[{Arnab} et~al.(2018){Arnab}, {Zheng}, {Jayasumana}, {Romera-Paredes},
  {Larsson}, {Kirillov}, {Savchynskyy}, {Rother}, {Kahl}, and
  {Torr}]{Arnab2018}
{Arnab}, A., {Zheng}, S., {Jayasumana}, S., {Romera-Paredes}, B., {Larsson},
  M., {Kirillov}, A., {Savchynskyy}, B., {Rother}, C., {Kahl}, F., and {Torr},
  P. H.~S.
\newblock Conditional random fields meet deep neural networks for semantic
  segmentation: Combining probabilistic graphical models with deep learning for
  structured prediction.
\newblock \emph{IEEE Signal Processing Magazine}, 35\penalty0 (1):\penalty0
  37--52, 2018.

\bibitem[Ashburner et~al.(2000)Ashburner, Ball, Blake, Botstein, Butler,
  Cherry, Davis, Dolinski, Dwight, Eppig, et~al.]{ashburner2000gene}
Ashburner, M., Ball, C.~A., Blake, J.~A., Botstein, D., Butler, H., Cherry,
  J.~M., Davis, A.~P., Dolinski, K., Dwight, S.~S., Eppig, J.~T., et~al.
\newblock Gene ontology: tool for the unification of biology.
\newblock \emph{Nature genetics}, 25\penalty0 (1):\penalty0 25--29, 2000.

\bibitem[Battaglia et~al.(2018)Battaglia, Hamrick, Bapst, Sanchez-Gonzalez,
  Zambaldi, Malinowski, Tacchetti, Raposo, Santoro, Faulkner,
  et~al.]{battaglia2018relational}
Battaglia, P.~W., Hamrick, J.~B., Bapst, V., Sanchez-Gonzalez, A., Zambaldi,
  V., Malinowski, M., Tacchetti, A., Raposo, D., Santoro, A., Faulkner, R.,
  et~al.
\newblock Relational inductive biases, deep learning, and graph networks.
\newblock \emph{arXiv preprint arXiv:1806.01261}, 2018.

\bibitem[Belanger \& McCallum(2016)Belanger and
  McCallum]{Belanger2016StructuredPE}
Belanger, D. and McCallum, A.
\newblock Structured prediction energy networks.
\newblock In \emph{ICML}, 2016.

\bibitem[Bronstein et~al.(2017)Bronstein, Bruna, LeCun, Szlam, and
  Vandergheynst]{bronstein2017geometric}
Bronstein, M.~M., Bruna, J., LeCun, Y., Szlam, A., and Vandergheynst, P.
\newblock Geometric deep learning: going beyond euclidean data.
\newblock \emph{IEEE Signal Processing Magazine}, 34\penalty0 (4):\penalty0
  18--42, 2017.

\bibitem[Bruna et~al.(2013)Bruna, Zaremba, Szlam, and LeCun]{bruna2013spectral}
Bruna, J., Zaremba, W., Szlam, A., and LeCun, Y.
\newblock Spectral networks and locally connected networks on graphs.
\newblock \emph{arXiv preprint arXiv:1312.6203}, 2013.

\bibitem[Corso et~al.(2020)Corso, Cavalleri, Beaini, Li{\`o}, and
  Veli{\v{c}}kovi{\'c}]{corso2020principal}
Corso, G., Cavalleri, L., Beaini, D., Li{\`o}, P., and Veli{\v{c}}kovi{\'c}, P.
\newblock Principal neighbourhood aggregation for graph nets.
\newblock \emph{arXiv preprint arXiv:2004.05718}, 2020.

\bibitem[Cuong et~al.(2014)Cuong, Ye, Lee, and Chieu]{cuong2014conditional}
Cuong, N.~V., Ye, N., Lee, W.~S., and Chieu, H.~L.
\newblock Conditional random field with high-order dependencies for sequence
  labeling and segmentation.
\newblock \emph{Journal of Machine Learning Research}, 15:\penalty0 981--1009,
  2014.

\bibitem[Deac et~al.(2019{\natexlab{a}})Deac, Huang, Veli{\v{c}}kovi{\'c},
  Li{\`o}, and Tang]{deac2019drug}
Deac, A., Huang, Y.-H., Veli{\v{c}}kovi{\'c}, P., Li{\`o}, P., and Tang, J.
\newblock Drug-drug adverse effect prediction with graph co-attention.
\newblock \emph{arXiv preprint arXiv:1905.00534}, 2019{\natexlab{a}}.

\bibitem[Deac et~al.(2019{\natexlab{b}})Deac, Veli{\v{c}}kovi{\'c}, and
  Sormanni]{deac2019attentive}
Deac, A., Veli{\v{c}}kovi{\'c}, P., and Sormanni, P.
\newblock Attentive cross-modal paratope prediction.
\newblock \emph{Journal of Computational Biology}, 26\penalty0 (6):\penalty0
  536--545, 2019{\natexlab{b}}.

\bibitem[Djuric et~al.(2011)Djuric, Radosavljevic, Coric, and
  Vucetic]{djuric2011}
Djuric, N., Radosavljevic, V., Coric, V., and Vucetic, S.
\newblock Travel speed forecasting by means of continuous conditional random
  fields.
\newblock \emph{Transportation Research Record: Journal of the Transportation
  Research Board}, 2263:\penalty0 131--139, 12 2011.
\newblock \doi{10.3141/2263-15}.

\bibitem[{Djuric} et~al.(2015){Djuric}, {Radosavljevic}, {Obradovic}, and
  {Vucetic}]{djuric2015}
{Djuric}, N., {Radosavljevic}, V., {Obradovic}, Z., and {Vucetic}, S.
\newblock Gaussian conditional random fields for aggregation of operational
  aerosol retrievals.
\newblock \emph{IEEE Geoscience and Remote Sensing Letters}, 12\penalty0
  (4):\penalty0 761--765, 2015.

\bibitem[Gao et~al.(2019)Gao, Pei, and Huang]{gao2019conditional}
Gao, H., Pei, J., and Huang, H.
\newblock Conditional random field enhanced graph convolutional neural
  networks.
\newblock In \emph{Proceedings of the 25th ACM SIGKDD International Conference
  on Knowledge Discovery \& Data Mining}, pp.\  276--284, 2019.

\bibitem[Gilmer et~al.(2017)Gilmer, Schoenholz, Riley, Vinyals, and
  Dahl]{gilmer2017neural}
Gilmer, J., Schoenholz, S.~S., Riley, P.~F., Vinyals, O., and Dahl, G.~E.
\newblock Neural message passing for quantum chemistry.
\newblock In \emph{Proceedings of the 34th International Conference on Machine
  Learning-Volume 70}, pp.\  1263--1272. JMLR. org, 2017.

\bibitem[Gligorijevic et~al.(2019)Gligorijevic, Renfrew, Kosciolek, Leman, Cho,
  Vatanen, Berenberg, Taylor, Fisk, Xavier, et~al.]{gligorijevic2019structure}
Gligorijevic, V., Renfrew, P.~D., Kosciolek, T., Leman, J.~K., Cho, K.,
  Vatanen, T., Berenberg, D., Taylor, B.~C., Fisk, I.~M., Xavier, R.~J., et~al.
\newblock Structure-based function prediction using graph convolutional
  networks.
\newblock \emph{bioRxiv}, pp.\  786236, 2019.

\bibitem[Hamilton et~al.(2017{\natexlab{a}})Hamilton, Ying, and
  Leskovec]{hamilton2017inductive}
Hamilton, W., Ying, Z., and Leskovec, J.
\newblock Inductive representation learning on large graphs.
\newblock In \emph{Advances in neural information processing systems}, pp.\
  1024--1034, 2017{\natexlab{a}}.

\bibitem[Hamilton et~al.(2017{\natexlab{b}})Hamilton, Ying, and
  Leskovec]{hamilton2017representation}
Hamilton, W.~L., Ying, R., and Leskovec, J.
\newblock Representation learning on graphs: Methods and applications.
\newblock \emph{arXiv preprint arXiv:1709.05584}, 2017{\natexlab{b}}.

\bibitem[Kalchbrenner et~al.(2016)Kalchbrenner, Espeholt, Simonyan, Oord,
  Graves, and Kavukcuoglu]{kalchbrenner2016neural}
Kalchbrenner, N., Espeholt, L., Simonyan, K., Oord, A. v.~d., Graves, A., and
  Kavukcuoglu, K.
\newblock Neural machine translation in linear time.
\newblock \emph{arXiv preprint arXiv:1610.10099}, 2016.

\bibitem[Kingma \& Ba(2014)Kingma and Ba]{kingma2014adam}
Kingma, D.~P. and Ba, J.
\newblock Adam: A method for stochastic optimization.
\newblock \emph{arXiv preprint arXiv:1412.6980}, 2014.

\bibitem[Kipf \& Welling(2016)Kipf and Welling]{kipf2016semi}
Kipf, T.~N. and Welling, M.
\newblock Semi-supervised classification with graph convolutional networks.
\newblock \emph{arXiv preprint arXiv:1609.02907}, 2016.

\bibitem[Kr{\"a}henb{\"u}hl \& Koltun(2011)Kr{\"a}henb{\"u}hl and
  Koltun]{Krhenbhl2011EfficientII}
Kr{\"a}henb{\"u}hl, P. and Koltun, V.
\newblock Efficient inference in fully connected crfs with gaussian edge
  potentials.
\newblock In \emph{NIPS}, 2011.

\bibitem[Lafferty et~al.(2001)Lafferty, Mccallum, and Pereira]{laferty2001crf}
Lafferty, J., Mccallum, A., and Pereira, F.
\newblock Conditional random fields: Probabilistic models for segmenting and
  labeling sequence data.
\newblock pp.\  282--289, 01 2001.

\bibitem[Liu et~al.(2019)Liu, Zhou, Long, Jiang, and Zhang]{liu2019learning}
Liu, L., Zhou, T., Long, G., Jiang, J., and Zhang, C.
\newblock Learning to propagate for graph meta-learning.
\newblock In \emph{Advances in Neural Information Processing Systems}, pp.\
  1037--1048, 2019.

\bibitem[Ma et~al.(2018)Ma, Xiao, Shang, and Sun]{ma2018cgnf}
Ma, T., Xiao, C., Shang, J., and Sun, J.
\newblock Cgnf: Conditional graph neural fields.
\newblock 2018.

\bibitem[Oord et~al.(2016)Oord, Dieleman, Zen, Simonyan, Vinyals, Graves,
  Kalchbrenner, Senior, and Kavukcuoglu]{oord2016wavenet}
Oord, A. v.~d., Dieleman, S., Zen, H., Simonyan, K., Vinyals, O., Graves, A.,
  Kalchbrenner, N., Senior, A., and Kavukcuoglu, K.
\newblock Wavenet: A generative model for raw audio.
\newblock \emph{arXiv preprint arXiv:1609.03499}, 2016.

\bibitem[Radosavljevic et~al.(2010)Radosavljevic, Vucetic, and
  Obradovic]{radosavljevic2010}
Radosavljevic, V., Vucetic, S., and Obradovic, Z.
\newblock Continuous conditional random fields for regression in remote
  sensing.
\newblock volume 215, pp.\  809--814, 01 2010.
\newblock \doi{10.3233/978-1-60750-606-5-809}.

\bibitem[Scarselli et~al.(2008)Scarselli, Gori, Tsoi, Hagenbuchner, and
  Monfardini]{scarselli2008graph}
Scarselli, F., Gori, M., Tsoi, A.~C., Hagenbuchner, M., and Monfardini, G.
\newblock The graph neural network model.
\newblock \emph{IEEE Transactions on Neural Networks}, 20\penalty0
  (1):\penalty0 61--80, 2008.

\bibitem[Snell et~al.(2017)Snell, Swersky, and Zemel]{snell2017prototypical}
Snell, J., Swersky, K., and Zemel, R.
\newblock Prototypical networks for few-shot learning.
\newblock In \emph{Advances in neural information processing systems}, pp.\
  4077--4087, 2017.

\bibitem[Springenberg et~al.(2014)Springenberg, Dosovitskiy, Brox, and
  Riedmiller]{springenberg2014striving}
Springenberg, J.~T., Dosovitskiy, A., Brox, T., and Riedmiller, M.
\newblock Striving for simplicity: The all convolutional net.
\newblock \emph{arXiv preprint arXiv:1412.6806}, 2014.

\bibitem[Veli{\v{c}}kovi{\'c} et~al.(2017)Veli{\v{c}}kovi{\'c}, Cucurull,
  Casanova, Romero, Lio, and Bengio]{velivckovic2017graph}
Veli{\v{c}}kovi{\'c}, P., Cucurull, G., Casanova, A., Romero, A., Lio, P., and
  Bengio, Y.
\newblock Graph attention networks.
\newblock \emph{arXiv preprint arXiv:1710.10903}, 2017.

\bibitem[Veli{\v{c}}kovi{\'c} et~al.(2019)Veli{\v{c}}kovi{\'c}, Ying, Padovano,
  Hadsell, and Blundell]{velivckovic2019neural}
Veli{\v{c}}kovi{\'c}, P., Ying, R., Padovano, M., Hadsell, R., and Blundell, C.
\newblock Neural execution of graph algorithms.
\newblock \emph{arXiv preprint arXiv:1910.10593}, 2019.

\bibitem[Xu et~al.(2018)Xu, Hu, Leskovec, and Jegelka]{xu2018powerful}
Xu, K., Hu, W., Leskovec, J., and Jegelka, S.
\newblock How powerful are graph neural networks?
\newblock \emph{arXiv preprint arXiv:1810.00826}, 2018.

\bibitem[Zhou et~al.(2019)Zhou, Jiang, Bergquist, Lee, Kacsoh, Crocker, Lewis,
  Georghiou, Nguyen, Hamid, et~al.]{zhou2019cafa}
Zhou, N., Jiang, Y., Bergquist, T.~R., Lee, A.~J., Kacsoh, B.~Z., Crocker,
  A.~W., Lewis, K.~A., Georghiou, G., Nguyen, H.~N., Hamid, M.~N., et~al.
\newblock The cafa challenge reports improved protein function prediction and
  new functional annotations for hundreds of genes through experimental
  screens.
\newblock \emph{Genome biology}, 20\penalty0 (1):\penalty0 1--23, 2019.

\bibitem[Zitnik \& Leskovec(2017)Zitnik and Leskovec]{zitnik2017predicting}
Zitnik, M. and Leskovec, J.
\newblock Predicting multicellular function through multi-layer tissue
  networks.
\newblock \emph{Bioinformatics}, 33\penalty0 (14):\penalty0 i190--i198, 2017.

\bibitem[Zitnik et~al.(2018)Zitnik, Agrawal, and Leskovec]{zitnik2018modeling}
Zitnik, M., Agrawal, M., and Leskovec, J.
\newblock Modeling polypharmacy side effects with graph convolutional networks.
\newblock \emph{Bioinformatics}, 34\penalty0 (13):\penalty0 i457--i466, 2018.

\end{thebibliography}

\end{document}